# A Novel Approach for Multi-Agent Cooperative Pursuit to Capture Grouped Evaders

**Muhammad Zuhair Qadir. Songhao Piao. Haiyang Jiang. Mohammed El Habib Souidi**

## Abstract

An approach of mobile multi-agent pursuit, based on application of Self-Organizing Feature Map (SOFM) and along with that reinforcement learning based on Agent Group Role Membership Function (AGRMF) model is proposed. This method promotes dynamic organization of the pursuers' groups and also makes pursuers' group evader according to their desire based on SOFM and AGRMF techniques. This helps to overcome the shortcomings of the pursuers that they cannot fully reorganize when the goal is too independent in process of AGRMF models operation. Besides, we also discuss a new reward function. After the formation of the group, reinforcement learning is applied to get the optimal solution for each agent. The results of each step in capturing process will finally affect the AGR membership function to speed up the convergence of the competitive neural network. The experiments result show that this approach is more effective for the mobile agents to capture evaders.

## 1. Introduction

Multi-agent systems can work together to solve different complex problems in a given environment. These agents can take the shape of a robot and can help in various situations. In the mid-1940s, William Grey Walter., in the study of the world's first artificial intelligent robot - the tortoise, found that these simple robots could perform some actions called "complex social behaviour" cooperatively [1]. Since then, significant progress has been made in the aspect theory and application of multi-agent systems. Pursuit is a rivalry between two groups/agents one being the pursuer and other being the evader. Multi-agent cooperative pursuit evasion has been a popular research area when it comes to games and other related problems [20]. Game theory have provided one of the simplest approaches to help solve this problem as discussed in previous literature [17]. It uses the value and reward system to guide the agents towards its goal in an efficient way. Satisfying the end goal of every agent. The benefit behind using game theory in MAS is that it provides coordination among them through mutual rationality of the multi-agents [21]. Moreover, Amigoni et al. [7] proposed a way to compute the optimal pursuer's strategy that maximizes chances of target's capturing in an environment. The method is based on theoretical model and mathematical modelling. In [6], author highlights the necessary conditions and optimal pursuit strategy for successful chase by surrounding the evaders based on a mathematical model and geometrical principles.

   Cooperation was considered an important aspect in a quicker pursuit in [5] where authors proposed a strategy based on pursuit cooperation using association rule of data mining technology. It was discussed in many other problems like task allocation for multi-

**Muhammad Zuhair Qadir**
*mzuhairqadir@gmail.com, Department of Computer Science and Technology, Harbin Institute of Technology, Harbin, China.*
**Songhao Piao (corresponding author)**
*piaosh@hit.edu.cn, Department of Computer Science and Technology, Harbin Institute of Technology, Harbin, China.*
**Haiyang Jiang**
*jhy786839364@163.com, Department of Mathematics, Harbin Institute of Technology, Harbin, China.*
**Mohammed El Habib Souidi**
*mohamed.souidi@hit.edu.cn, Department of Computer Science, University of Khenchela, Khenchela, Algeria.*

**Funding: This paper is supported by National Natural Science Foundation of China [grant number 61375081]; a special fund project of Harbin science and technology innovation talents research [grant number RC2013XK010002].**

agents to make them cooperate on related tasks just as in [22]. Cooperation helps in robust and quick goal achievement. An interesting example of cooperation in task completion is discussed in table cleaning [25]. It shows the importance of task allocation in cooperative way so that performable tasks and agent capabilities are better utilized and goal gets easier to be obtained by working together. In another study, Durham et al. [24] discussed how to make a strategy to handle clearing problem in which agents cooperatively guarantee to clear an environment given enough numbers of searchers are available. SOFM also known as Kohonen map being one of the most popular in artificial neural networks uses competitive learning to make a map and cluster objects together. It works similar to K-means but with the difference that vectors which are closer in high-dimensional space also end up being mapped to nodes that are close in two dimensional space. Artificially intelligent agent uses the Q-learning method to learn from their interactions and help in pursuit evasion problem as presented in [23] but they work independently and didn't consider the group formation for the task in hand.

With these current research, we were able inspire and propose our own algorithm which can better handle the problem with the help of reorganization to improve capture time. In section 3, we try to explain the pursuit evasion problem, moving on to discussing the mathematical model and our algorithm working in subsequent sections.

## 2. Related Work

Cooperative pursuit evasion is a known multi-agent problem which has attracted a lot of attention in the field of multi-agent robotics with a wide variety of applications such as search and rescue in disaster area and patrol on a larger scale etc. to name a few. It also has applications in home care, military combat and mobile sensor networks. For example cooperative bodyguards can be of good use [26] [27]. Pursuit evasion is widely researched in differentials games as well [15] [16]. There are two key points in the implementation of coordination of multi-agents pursuit, one is organization and the other one is reorganization [2]. Organization means that each pursuer can find its own task by itself or just be assigned by control system according to its group's situation, all of their social behaviour can be determined by their own ability and condition, the task of the whole group and the role played in the group. Reorganization is a capability of multi-agent system which promotes itself to move to a different organization which will be more adaptive to the new environment and hence can make the process of executing the task more effective than a previous organization. Reorganization is also referred as flexibility. There has been many useful and novel approaches applied to improve cooperative multi-agent systems (MAS), in which some researchers focus on interaction between the agents and the environment such as graph theory [3], polygonal environment [4], cellular environment [5] and some others researchers focus on mechanisms of cooperation among human individuals and tries to find out different ways to apply multi agent pursuit effectively. For instance, Awang Chen et al. proposed a new hunting strategy based on optimized quick surrounding and quick-capture direction [6]. Amigoni et al. came out with a game based theoretical approach to find optimal strategies for pursuit-evasion in grid environments [7]. In order to simplify the process of communication of agents, Ze-Su C. et al. ameliorated a kind of task negotiation process based on Multi-Unit Combinatorial Auction [8]. Hao Wang et al. presented an alliance generation algorithm which can determine the cooperation intention according to the agent's emotional factors, so that pursuers with the negative emotions can be prevented from participating in the group in case of a negative impact on the alliance [9]. Mohammed El Habib Souidi et al. proposed an approach which can generate agents' membership function called AGRMF which consists of their own credit and confidence just like human being, with the result that MAS can get optimized collaboration in grid environment making use of fuzzy logic [10].

These methods and research have dramatically improved the state-of-the-art of multi-agent cooperation pursuit. One of the most important problems that we are facing in this field is that pursuers can't fully reorganize when evaders are too independent. This paper addresses this problem by proposing an algorithm for mobile Multi-agent pursuit based on application of SOFM and Reinforcement learning on AGR model, this method promotes dynamic organization of the pursuers groups and also makes the pursuers group evaders according to their will based on SOFM and AGRMF. It also helps the pursuers' to be more dispersed and spread out as the process of chase goes on. In addition, a new reward function is discussed. Once the formation of the group is made, reinforcement learning technique is applied to get the optimal solution for each agent including the evaders. After each step capture process becomes easier as it will affect the AGR membership function and also speed up the convergence of the competitive neural network. Finally, we analyze the effectiveness of the algorithm from the macro and micro simulation results.

## 3. Description of multi - agent pursuit problem

### *3.1. Problem description of pursuit-evasion*

The multi-agent systems for pursuit-evasion problem based on AGR model mainly consist of two types of agents, pursuers and evaders .The set of pursuers is denoted by P and the set of evaders is denoted by E: $P = \{p_1, p_2, \dots p_n\}$ And $E = \{e_1, e_2 \dots e_n\}$. One of the features of the AGRMF model is that it gives pursuers two parameters that describe their capabilities learned from the process of pursuit, one is confidence $Conf$: $\forall Conf \in [0.1, 1]: max\left(0.1, \frac{c_s}{c_t}\right)$ Another one is Credit $Cred$. $\forall Credit \in [0.1, 1]: min\left(1, 1 - \frac{c_b}{c_t + c_s}\right)$. Here $C_s$ is the number of tasks that has been accomplished successfully, $C_t$ is the number of tasks in which the agent has participated. $C_b$ is the number of the evaders abandoned by the agent. Each pursuer has its own pursuit range which is denoted by set $R$: $R = \{r_1, r_2, ... r_n\}$, the evaders which come into $r_n$ will be given priority to by $p_n$. One of the pursuers will be selected to be the organizer, which will assign the pursuers to groups and cluster similar evaders to the same group according to AGRMF.    Pursuit difficulty means that how many pursuers are needed to catch that evader, the set of pursuit difficulty is denoted by D: $= \{d_1, d_2, ... d_n\}$ . Besides, each evader has their own priority, which is dependent on how many pursuers are near to that evader and secondly, inversely proportional to difficulty for catching that evader. So more pursuers close to that evader and lower the difficulty, the higher will be the priority assigned to that evader. Priority of an evader is denoted $pr_i$:

$$pr_i = \frac{1+P(e_i)}{d_i} \qquad (1)$$

Here *P(e_i)* represents the number of pursuers whose pursuit range is invaded by evader *e_i*. One group consists of one or more pursuers and evaders and the pursuers will try to pursue the evaders in that group. One evader will be considered to have been arrested by pursuers and dispersed from environment if there are enough pursuers or obstacles around him blocking his way as equation 2:

$$d_n \leq \sum_{x \in neibor(d_n)} P(x) + O(x) \qquad (2)$$

Here *neighbour(d_n)* means the adjacent cell. *P(x)* means the grid cell *x* is whether occupied by pursuers or not and *O(x)* means whether that is an obstacle or not.

## *3.2. Agent group role membership function*

Agent group role membership function (AGRMF) of pursuer *P* to *E* after the modification based on the expression proposed in literature [4] is shown in equation 3 as:

$$\mu_e(p) = \frac{Coef_1 * Dist_{e,p} + Coef_2 * Conf + Coef_3 * Credit}{\sum_{i=1}^{3} Coef_i} \qquad (3)$$

Where $Dist_{e,p}$ means Cartesian distance between *e* and *p*. Ordinary $\mu_e(p)$ represents connection between pursuer *p* and evader *e*. *Coef1 to Coef3* are the main ability indicators of pursuer *p* which are used to determine whether it can play a role in capturing *e*. The basic algorithm for coalition formation tries to greedily select enough pursuers with the highest $\mu_e(p)$ to join the group containing *e* based on AGRMF [4].

## **4. Grouping evaders and path planning**

The ordinary AGR model as extended with AGRMF creates one group for each evader which also means one group consists only one evader and the pursuers' group only focus on that evader letting alone others. This is the reason why the pursuers and evaders are relatively more discrete therefore the self-organization becomes less flexible when there are more evaders and in the bigger grid environment. This situation is shown in section 5. This subsection discusses a novel approach based on AGRMF which applies SOFM to generate one group consisting of more than one evaders.

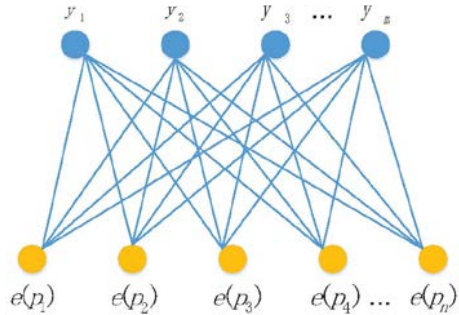

*Figure. 1: Structure of SOFM driven by AGRMF*

The vector of AGRMF of evader n is denoted by $U_{e_n} = \{U_{e_n}(p_1), U_{e_n}(p_2), U_{e_n}(p_3) ... U_{e_n}(p_m)\}$. The set of $U_{e_n}$ contains the chosen representation of evaders and are fed to SOFM to promote and learn how to group similar evaders. The basic structure of SOFM is inspired by AGRMF is shown in *Figure 1*. The proposed algorithm we have used for training SOFM are also inspired by AGRMF and through this we get final group as an output. The algorithm works by first initializing the parameters and SOFM then the learning process of SOFM begins and is inspired by [11] [12]. Finally, the output is the indexes of groups, each evader belongs to. The evader who gets the same output from SOFM will be assigned to the same group which will be discussed in detail in section 5.

## 4.1. Path planning of multi–agent

Once the various pursuer alliances are formed, the next step is to determine the plan of motion. For this, we apply reinforcement learning as an agents' motion strategy. Reinforcement learning is a classical approach which is famous for solving motion optimization problem in the environment where agents can get delayed reward. Nowadays, application of reinforcement learning have improved the state-of-the-art of artificial intelligence such as unmanned cars, unmanned airplane and deep reinforcement learning [15]. In this method, we developed a novel reward function to transform the theoretical model to its application in the mobile agent's field.

In reinforcement learning, agents can perceive the different states of the environment set $S$, and the different actions they can execute set $A$. For every discrete step, agents perceive the current state $s_t$ and selects an action $a_t$ to execute it. Environment responds to this agent and returns reward $r_t = (s_t, a_t)$ here $r$ is an immediate payoff function and a successor state $s_{t+1} = \sigma(s_t, a)$ is then generated.

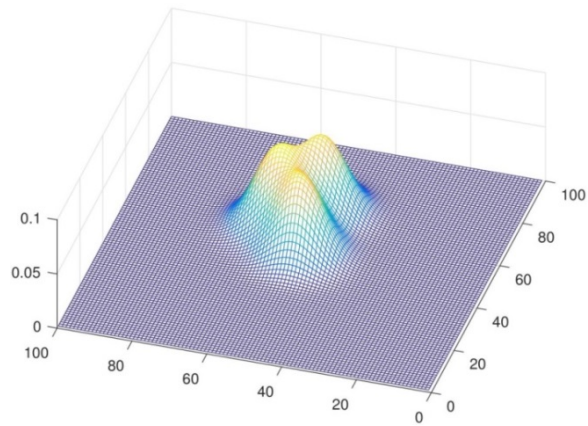

Figure. 2: Immediate payoff function

## 4.2. Immediate payoff function

The $Re$ of evaders which determines the reward for the pursuers having caught it can get is equivalent to $D$, $Re_i$ is the reward in a certain region subject to normal distribution instead of occupying only one cell. The immediate payoff function of each cell which one pursuer gets from the organizer is sum of $Re$ of evaders of same group which that pursuer belongs to. The immediate pay off function of group is shown as equation 4 as below:

$$r_{group}(x,y) = \sum_{i \in group} Re_i pr_i e^{-\frac{1}{2}(x-x_i)^2 + (y-y_i)^2} \tag{4}$$

where $(x_i, y_i)$ is the Cartesian coordinates of the $e_i$. This kind of immediate reward function can increase the attraction of the pursuers towards the evader along with its neighbours and centre of the group just as shown in *Figure . 2*. Peaks in the graph shows the overlapping region of the evaders, targeting a region where the chances of capturing is increased and helps surround the group effectively. The pursuers prefer to surround all of them instead of focusing on only one of them, making it a very effective way to capture them all in groups.

## 4.3. Optimal strategy

Agent's task is to learn a strategy $\pi: s \to a$ which selects the next action $a_t$ based on the current state $s_t$, and enables the agent to get the maximum of discounted cumulative reward $V^\pi = \sum_{i=0}^{\infty} \lambda^i r_{t+i}$ where $\lambda$ is the discount factor, agents uses the principle of Q learning to complete the task, the function of agents getting to their next action is shown in algorithm 1. In which from step 2 to step 12, $Q(s,a)$ is continuously changing until all $Q(s,a)$ becomes optimal.

## 5. Working of our algorithm

This section shows the sequences of our algorithm to generate coalitions and execute the chase which is inspired by the strategies and functions based on previous sections. This algorithm flowchart is shown in *figure 4* is interpreted as following: The organizer evaluates the parameters including position and reward of each evader found in the environment by all pursuers and put them in list of evaders, then it broadcasts the message about each evader agent to all of the pursuers and wait for the response from pursuers. The pursuers sends the response to organizer after getting inquiry. Next, the organizer calculates the membership degree of each pursuer for each evader. The evaders with same output from the SOFM which is fed by their AGRMF vectors are divided into the same group, *pursuergroup* stores the list of pursuers which are responsible for each evader sequentially. Then the organizer will append all the pursuers who are ready to catch the evaders of the same group to the same coalition. Each alliance has a limited lifetime called *life* to avoid infinite catching situation and produce chances for reorganization, for each pursuer to follow the optimized strategy they calculate according to the immediate payoff matrix during that period. All of the pursuers' capability will result as a reward if one evader of the same group is caught. On the contrary, they will be punished for the rest of the evaders after lifetime. This whole scenario will continue until all evaders are captured.

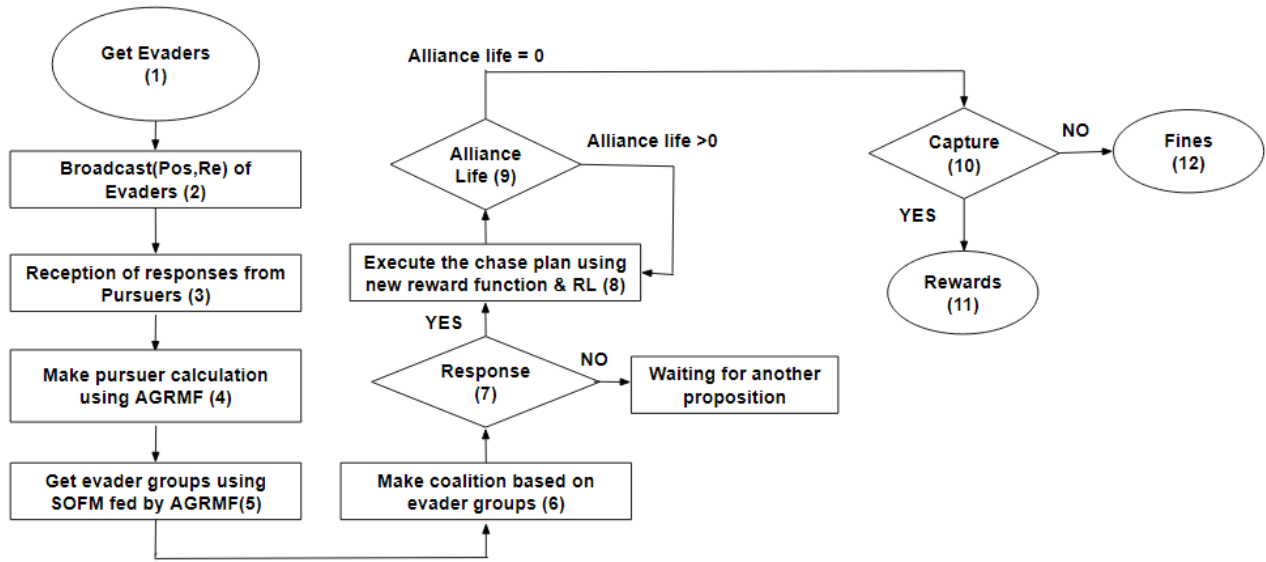

Figure. 4. Flowchart of the algorithm

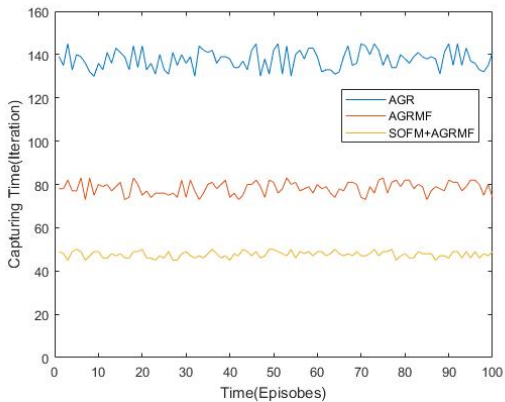
Figure 5. Average capturing time.

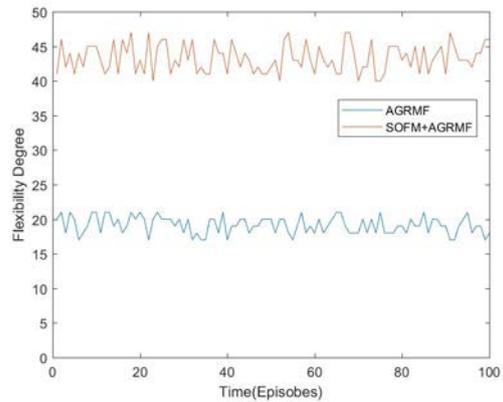
Figure 6. Average flexibility degree.

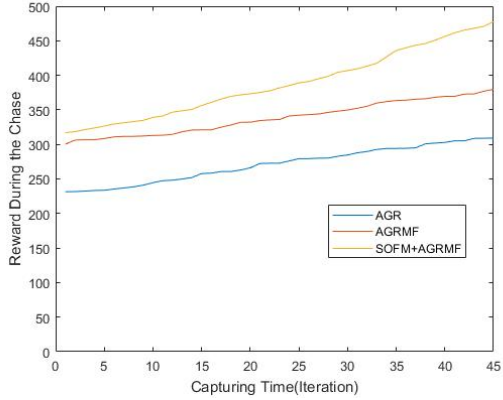
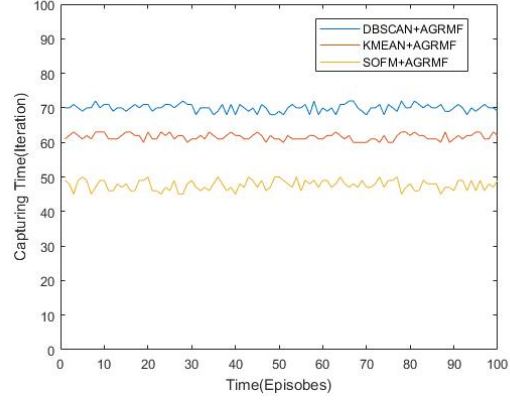

Figure 7. Development of Reward    Figure. 8. Capture time of different approaches for clustering evaders

## 6. Simulation experiments

We did experiments to show and reveal that how important is cluster similar evaders. We also compare the performance of our algorithm with other famous techniques. We implemented and experimented using grid environment which is widely used by scientific community to test and prove the effectiveness of the multi-agent systems [3] [4] [7]. Real world can easily be mapped into grid environment by using small units. In our case, we use the grid environment to carry out these simulations is a rectangular field of *100 by 100 grids*, in which there are *33 pursuers* and *9 evaders* whose *difficulty* vary between *2* and *4*. The action set of each agent is defined as $A = \{a_{up}, a_{down}, a_{right}, a_{left}\}$. All of agents have the same speed (one grid for each iteration). After *100* experiments of pursuits, results show that the pursuers of AGRMF model inspired by SOFM prefer to get together and surround evaders which are closed to each other than that of AGRMF model without SOFM.

We did experiments on two perspectives, first two cases shows how formation of pursuers is done but ignores evader clustering and last three showcase how the evaders are clustered together as well. More details about the experiments performed in different cases and notable outcomes are mentioned below:

- Case AGR: formation of the groups of pursuers without any mechanism integration which is totally based on the AGR organizational model and ignore evader group formation.
- Case AGRMF: formation of the groups of pursuers with the application of only AGRMF before any new chase iteration thereby correcting membership function during the first chase iteration. Also ignoring evader group formation.
- Case SOFM+AGRMF: formation of the groups of pursuers with the application of our coalition algorithm based on AGRMF to cluster evaders as well.
- Case KMEAN+AGRMF: formation of the groups of pursuers with the application of KMEANS to cluster similarly evaders with the most suitable parameters among the dataset.
- Case DBSCAN+AGRMF: formation of the groups of pursuers with the application of DBSCAN to cluster similarly evaders with the most suitable parameters among the dataset.

The results shown in *figure 5* are calculated after 100 experiments for each case. In the first case of AGR, the average of total chase iterations needed for capturing all evaders is 137.71 .For the second case of AGRMF, capturing of evaders is completed in an average of 78.27 chase iterations and when it comes to case of SOFM, the capture of the evaders has only consumed an average of **47.64** chase iterations for capturing the evaders. Making it a huge improvement on previous techniques that's an average almost **39.1%** improvement over AGRMF time for capturing evaders.

We also did experiments to compare the flexibility degree which is defined as number of times the reorganization takes place once new formation is needed, in the case of AGRMF and case of SOFM+AGRMF. If at some point, one of the pursuer/s change their target, its fellow group members need to reorganize and make a new coalition to effectively capture the target evader. In other words, flexibility defines as how many times pursuer agents reorganize and make a coalition formation towards the evader agents, if one of the pursuer change the target evader. So higher the flexibility, the better is their ability to reorganize and hence helping in capturing effectively. The results are shown in *figure 6*. Our experiments show that our algorithm helps agents reorganize on average **43.54 times** as compared to only 19.1 times in case of AGRMF alone which is an average almost more than double the times as a consequence reducing capturing time as well. Orange graph line shows significant improvement over AGRMF used without SOFM shown in blue graph line. So it can reorganize more abruptly and more times than without SOFM. *Figure 7* shows the results of the development of pursuer's reward in case of AGRMF and the case of SOFM+AGRMF. More rewards for better coalition and capturing the evader as the capture process continues and resulting in better reward graph as it approaches towards claiming all the evaders.

On the other hand, the comparison experiments about different approaches of clustering evaders is also done to showcase the performance of our algorithm. We compared popular KMEANS+AGRMF and DBSCAN+AGRMF approaches with our approach [13] [14]. Results show that the algorithm inspired by SOFM performs by par with the previous methods as shown in *figure. 8*. Our algorithm SOFM+AGRMF takes an average of **47.64** iterations to capture all the evaders in comparison with the KMEANS taking 61.6 and the DBSCAN+AGRMF taking 69.98 iterations. Overall significant improvement of **22.66%** over KMEANS. Overall, our algorithm improved capturing time and flexibility degree with the help of new reward function and coalition formation technique which clusters similar evaders as well.

## 7. Conclusion

In this paper, the cooperation mechanism of multi-agent pursuers based on SOFM and AGRMF is presented specifically focused on how to cluster similar evaders according to their membership function instead of just assigning group for each pursuer, besides a novel immediate payoff function is proposed to increase the attraction of cell around the evaders. Results show that the pursuers of AGRMF model inspired by SOFM are more effective to get together and surround evaders which are closed to each other than that of AGRMF model without SOFM. An improvement of 39.1% is achieved to capture all evaders with this new coalition formation. The main purpose of this algorithm is to improve the flexibility degree – the number of times agents adjust their coalition formation w.r.t. target - when the pursuers are in the situation of bigger environment and also more evaders in that environment. Our algorithm reorganizes almost double the times as compared to AGRMF. The simulation results show great improvement in reducing the total capturing time and improving the flexibility degree as compared to previous algorithms which doesn't cluster the evaders at all. Also it shows that it is adaptive to the change of environment.

Our algorithms works very well when evaders are randomly placed in an environment but is not much effective when evaders are too independent for example are at corners of the environment and will take same time as AGRMF. In future, we would like to tackle this problem. Other aspects is to consider camera on agent as input with limited range of view and variable speed of agents in pursuit-evasion.